\documentclass[11pt]{article}

\usepackage{bm}
\usepackage[normalem]{ulem}
\usepackage{amsmath,graphics,epsfig,float,braket,multirow,url}
\usepackage{caption}
\usepackage{subcaption}
\usepackage[toc,page]{appendix}
\usepackage{amsmath}
\usepackage{microtype}
\usepackage{comment}
\usepackage{authblk}
\usepackage{hyperref}
\usepackage[capitalize]{cleveref}
\usepackage[normalem]{ulem}

\newcommand{\s}[1]{{\textsf{\textbf{#1}}}}

\begin{document}

%%%% Article title to be placed here
\title{\s{A passive universal grasping mechanism based on an everting shell}}
\author{ \textsf{Mythra V. S. Balakuntala $^\dagger$, Safvan Palathingal $^\dagger$, and G. K. Ananthasuresh $^\dagger$}}
\date{
{\it $^{\dagger}$Indian Institute of Science, Bengaluru, India}\\[2ex]
December 8, 2018
}

 \maketitle
\hrule\vskip 6pt

%%%% Abstract text to be placed here %%%%%%%%%%%%
\begin{abstract}
A passive monolithic compliant grasping mechanism that works based on the eversion of an elastically deformable bistable shell is conceptualized. It comprises grasping arms made of beam segments that work in conjunction with the everting shell. The grasper is capable of picking up a stiff object of any shape up to a maximum size and weight. The bistable shell everts upon contact with the object to enable the grasping arms envelop the object forming an enclosure. The mechanism then stays in that configuration until it is actuated again to turn the shell back to its original configuration and thereby opening the enclosure to release the object. The stiffness of the arms decides the payload of the mechanism. The size of the arms decides the largest object that can be grasped and held. The arms have distributed compliance so that they can conform to the shape of the object without applying undue force on it.

\end{abstract}

\vskip 6pt
\hrule
\vskip 6pt
\section{Introduction}
Universal passive grippers ought to grasp objects of varying size and shape. In comparison to active multi-fingered grippers, they do not need sophisticated grasping algorithms and thus are easier to implement \cite{bicchi2000robotic,murray2017mathematical}. Passive grippers can be broadly classified as follows: grippers with soft contacts \cite{choi2006design,maruyama2013delicate}, granular grippers \cite{amend2012positive, brown2010universal,manti2016stiffening}, and underactuated grippers \cite{laliberte2002underactuation,stavenuiter2017planar}. In this work, we explore an alternative way to passively grip objects of arbitrarily shaped objects using a monolithic compliant grasping mechanism based on bistable shells, also known as everting shells.

\begin{figure}
	\centering
	\includegraphics[width=\linewidth]{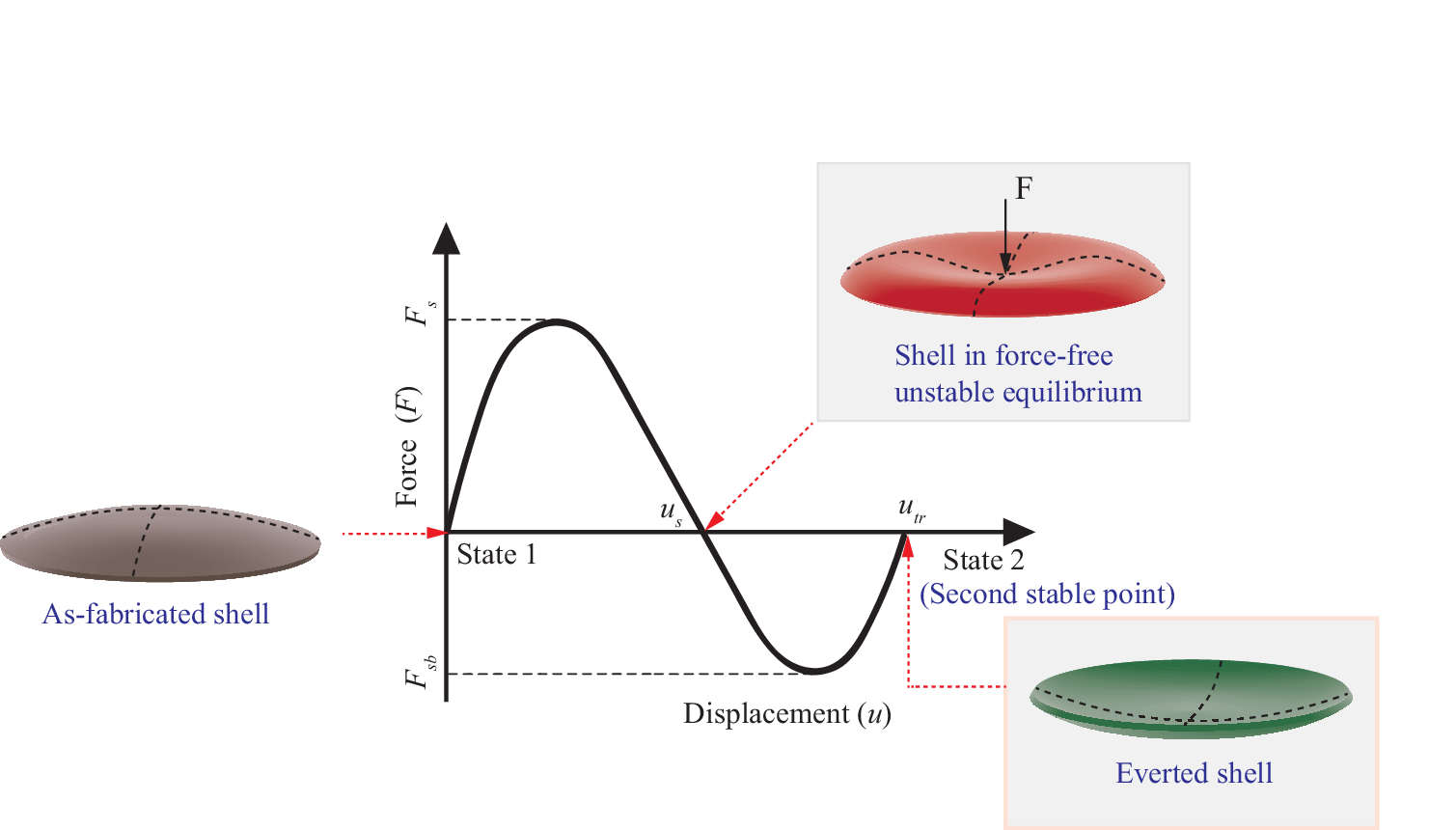}
	\caption{A typical force-displacement curve of a bistable shell.}
	\label{fig:typicalforcedelta}
\end{figure}

Everting shells can maintain two force-free equilibrium states as shown in the \cref{fig:typicalforcedelta}. The bistable shell considered here is stress-free in its as-fabricated state and stressed in its everted state. The ability to maintain two structural orientations without consuming power makes them ideal to be used in a passive gripper. A planar passive gripper using bistable arches was considered before by Nguyen and Wang \cite{nguyen2016gripper}. We extend this concept by using everting shell as the critical element and attaching grasping arms to it to conceive a gripper that is passive as well as capable of picking up objects of a variety of shapes.

The gripper is a monolithic design that combines a switching mechanism, an everting shell, and grasping arms. The switching mechanism releases grasped objects  by transmitting the input force to the everting shell. The bistable shell everts upon contact with the object to enable the grasping arms envelop the object. In Section 2, we explain the working principle of the gripper in detail. Furthermore, the analysis of the everting shell, switching mechanism and grasping arms is presented. We use finite element analysis (FEA) in ABAQUS \cite{hibbett1998abaqus} software to obtain force-displacement characteristics for the selected shape of the shell. The switching mechanism is designed by incorporating the force needed to evert the shell and maximum actuation force available.  The grasping arms are modeled as cantilever beams to obtain an approximate measure of payload. A 3D-printed prototype and results from preliminary grasping trials are given in Section 3. Section 4 includes the scope for future work.

\section{Design}
\subsection{Working principle}
\begin{figure}[h]
	\centering
	\includegraphics[width=0.9\linewidth]{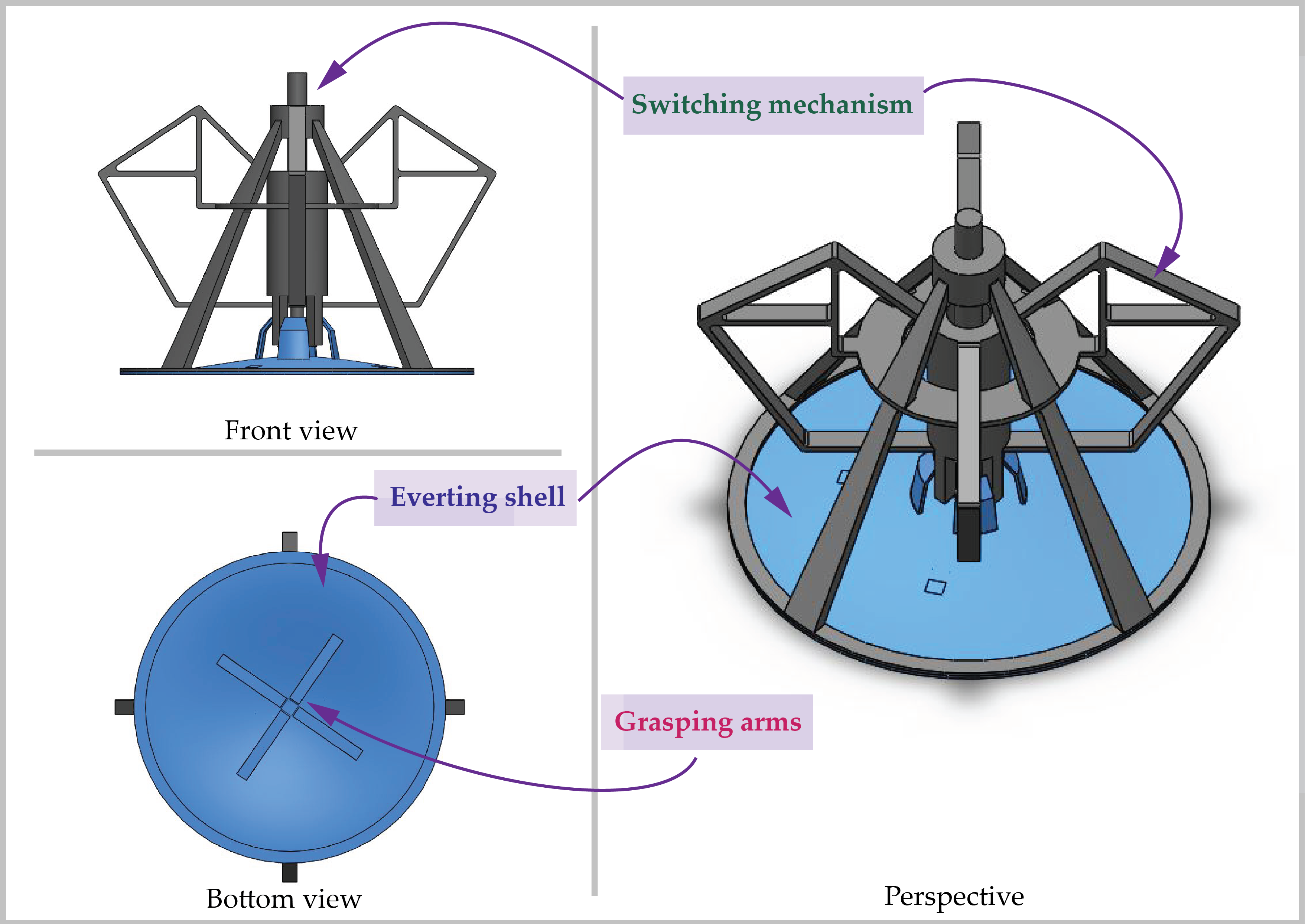}
	\caption{The compliant monolithic grasping mechanism based on an everting shell. The mechanism consists of three parts, (1) switching mechanism, (2) everting shell, and (3) grasping arms.}
	\label{fig:fullMechanism}
\end{figure}

\begin{figure}[b]
  	\centering
  	\includegraphics[width=0.9\linewidth]{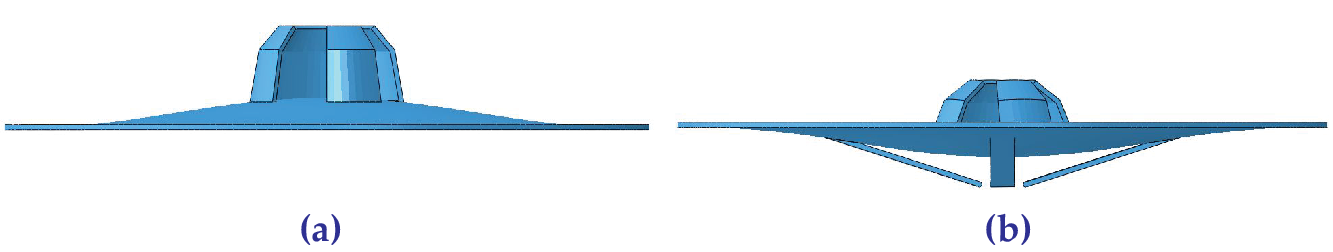}
  	\caption{Bistable shell with contracting mechanism and grasping arms attached.}
  	\label{fig:gripperSideView}
  \end{figure}

\begin{figure}[h]
 	\centering
 	\includegraphics[width=0.9\linewidth]{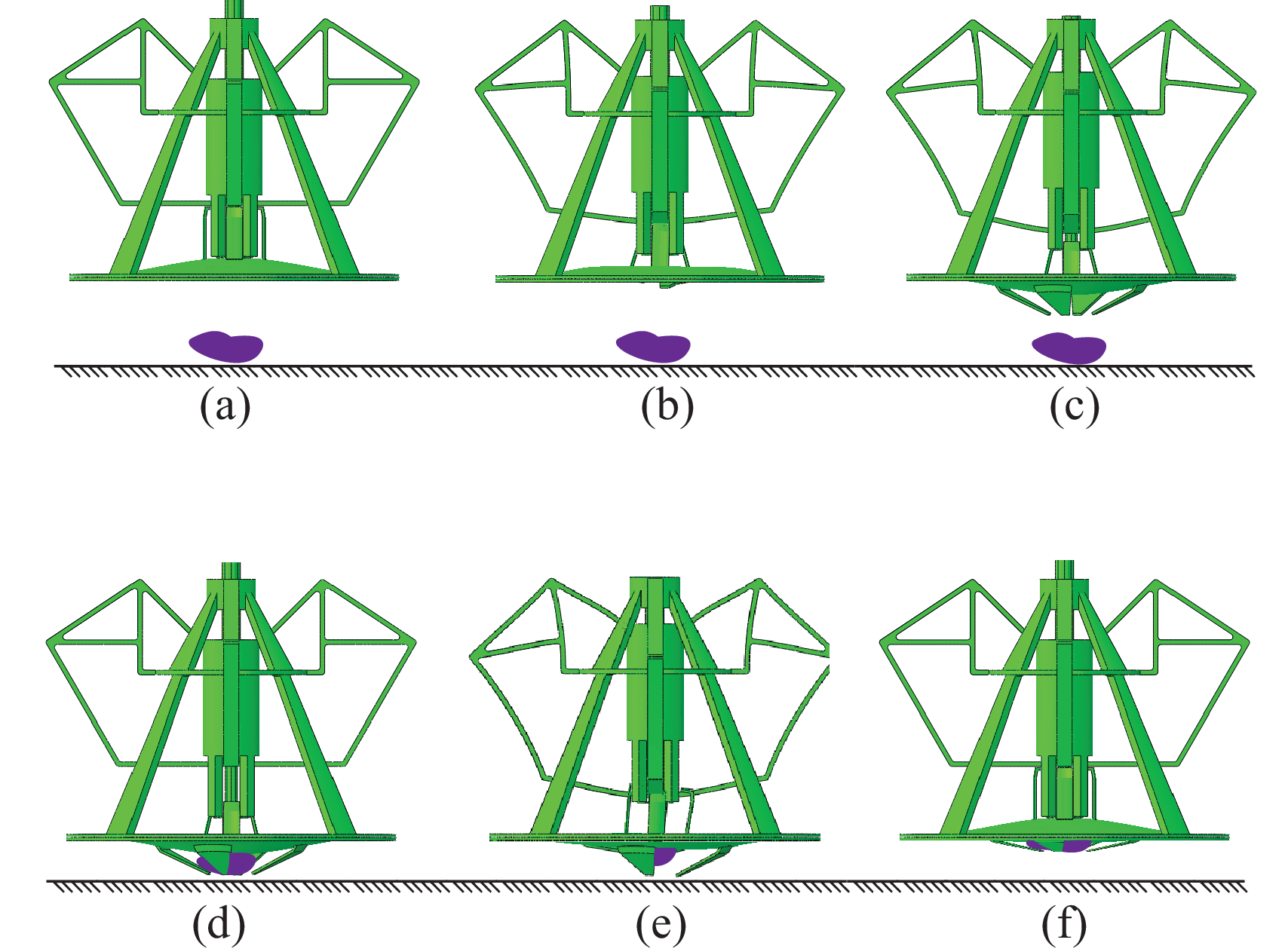}
 	\caption{Working of the grasping mechanism.}
 	\label{fig:timeLapse}
 \end{figure}

The everting compliant gripper consists of the switching mechanism, an everting shell, and grasping arms as shown in \cref{fig:fullMechanism}. 
 The switching mechanism is the connecting element between the actuator and the everting shell. It switches the everting shell from its stress-free configuration to the second stable everted configuration of the shell.
 \Cref{fig:gripperSideView} shows the grasping arms opening and closing in conjuction with the everting shell. 
As illustrated in \cref{fig:timeLapse}, these three parts work in the following three steps to grasp and release objects: (1) the input force applied on the switching mechanism switches the everting shell and open the grasping arms to grasp the object of interest (see \cref{fig:timeLapse}(a)-(c)); (2) the everted shell upon contact with the object returns  to the stress-free configuration causing the grasping arms to close around the object and grasp it  (see \cref{fig:timeLapse}(d)-(f)); (3) the switching mechanism is actuated again to release the object. 
 
 The design of the switching mechanism and the grasping arms is dependent on the bistable characteristics of the everting shell. These characteristics determine the topology and size of the switching mechanism, and the dimensions of the grasping arms as explained in the next two subsections. We systematically approach the analysis of the gripper considering the design of an everting shell first, followed by the switching mechanism and the grasping arms.

\subsection{Everting shell}

The two force-free stable equilibrium configurations of the everting shell correspond to the open and closed positions of the gripper. The critical static characteristics of the shell are switching ($F_s$) and switch-back ($F_{sb}$) forces, switching displacement ($u_s$), and the travel ($u_{tr}$). These characteristics are labeled in the force-displacement curve given in \cref{fig:typicalforcedelta}. $F_s$ is the minimum force required by the shell to switch from its stress-free stable configuration to the stressed everted shape. $F_{sb}$ is the maximum force that can be resisted by the shell in the everted state before switching back to its as-fabricated initial shape. $u_s$ is the minimum displacement the shell needs to be actuated to initiate its eversion and $u_{tr}$ is the total displacement of the point of interest between the two stable states. Note that the everting shell considered here is bistable solely because of its as-fabricated shape, not due to pre-stress. This is preferred as the gripper in its default closed configuration is stress-free.

\begin{figure}[h]
	\centering
	\includegraphics[width=0.9\linewidth]{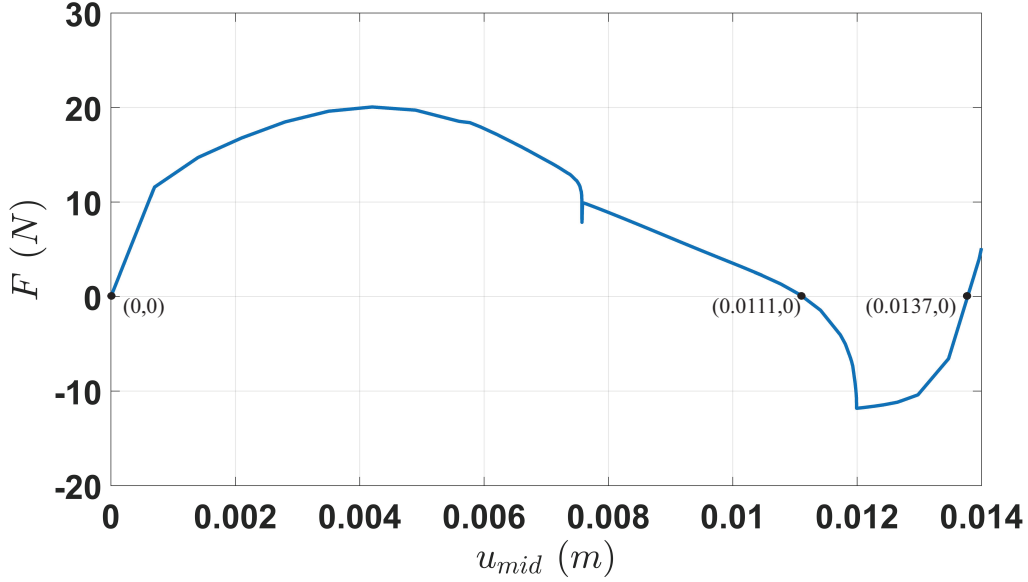}
	\caption{Bistable shell with contracting mechanism and grasping arms attached.}
	\label{fig:fdCosineShell}
\end{figure}

In the case of planar arches, it is known that a cosine profile does not show bistability for fixed-fixed boundary conditions \cite{palathingal2017design}. Here, we show that a shell formed from a revolved cosine profile is bistable for fixed-periphery boundary conditions. The height of the shell at a distance $r$ from the center, $h(r)$, is taken as:
\begin{equation}
	h(r)=\frac{h_{mid}}{2}\left[1-\cos\left(2\pi-\frac{2\pi r}{R}\right)\right]
\end{equation}
 where $h_{mid}$ is the height at the midpoint of the shell and $R$ the radius of the planform. We analyze the shell for an uniform thickness, $t=1~$mm, $h_{mid}=7~$mm, and $R=30~$mm using a quasi-static FEA in Abaqus. The dimensions were selected for realizing a prototype design that can be actuated by hand. Young's modulus and Poisson's ratio of Verowhite and TangloPlus mixture, a 3D printing material used in Objet Conex 260 for prototyping, were taken as $E=1.2~$GPa and $\nu=0.3$ respectively. For a point load applied at the center of the shell with edge fixed, the force displacement characteristics obtained is shown in \cref{fig:fdCosineShell}. Note that $u_{mid}$ is the deformation at the point of application of the load, $F$. The curve intersects the $u_{mid}$ axis, $F=0$, at three points corresponding to two stable and one in-between unstable states, indicating bistability.  
Two important design considerations for the prototype gripper, were sufficiently high $\frac{F_{sb}}{F_{s}}$ ratio and low $F_{s}$. An ideal $\frac{F_{sb}}{F_{s}}$ ratio of one implies that the forces required to switch from one state to the other are identical. $F_{s}$ is kept low so that the gripper can be actuated by hand. The dimensions of the shell were arrived at by considering multiple designs  such that $\frac{F_{sb}}{F_{s}}>0.5$ and $F_s<35~$N.

\subsection{Switching mechanism and grasping arms}

The switching mechanism transmits the force from the actuator to the center of the bistable shell. Hence, the parameters that affect the switching mechanism design are force and output displacement from the actuator, and switching force and displacement of the shell. For the everting shell considered here, $F_s=20N$ and $u_s=11.12~$mm. The switching mechanism selected here attaches to the edge of the shell as shown in \cref{fig:fullMechanism}. The force is applied at the protruding part at the top of the central axis of the mechanism. This initiates a contact at the top of the shell. When the deformation at this point of contact exceeds $u_s$, the shell everts to the second stable state. For  other cases where force or displacement amplification is required, the switching mechanism could be designed to be a force or displacement amplifying compliant mechanism without affecting the aforementioned grasping functionality \cite{krishnan2008evaluation}.

Four symmetrically arranged rectangular beams  that are attached to the bottom surface of the everting shell act as the grasping arms that hold objects. In the as-fabricated state of the arch, the grasping arms assume horizontal configuration. When the shell is in the second stable state, grasping arms protrudes outside as shown in \cref{fig:gripperSideView}. The grasping arms should be wide apart when the shell is in the second stable state. This is important to hold objects of larger size. This determines the length of the arms and the position at which they are attached to bottom surface of the shell. For the prototype considered here, beams of length $12~$mm attached at a distance of $15~$mm from the shell edge satisfy these constraints. When the everted shell comes in contact with stiff objects switches back to its initial state. In the process, the grasping arms come together and hold the object firmly. By taking the width of the beam as $5~$mm and depth as $1~$mm grasping are compliant enough wrap around the object, at the same time, stiff enough to support the weight of the object.

\section{Results and performance}
\begin{figure}[h]
	\centering
	\includegraphics[width=\linewidth]{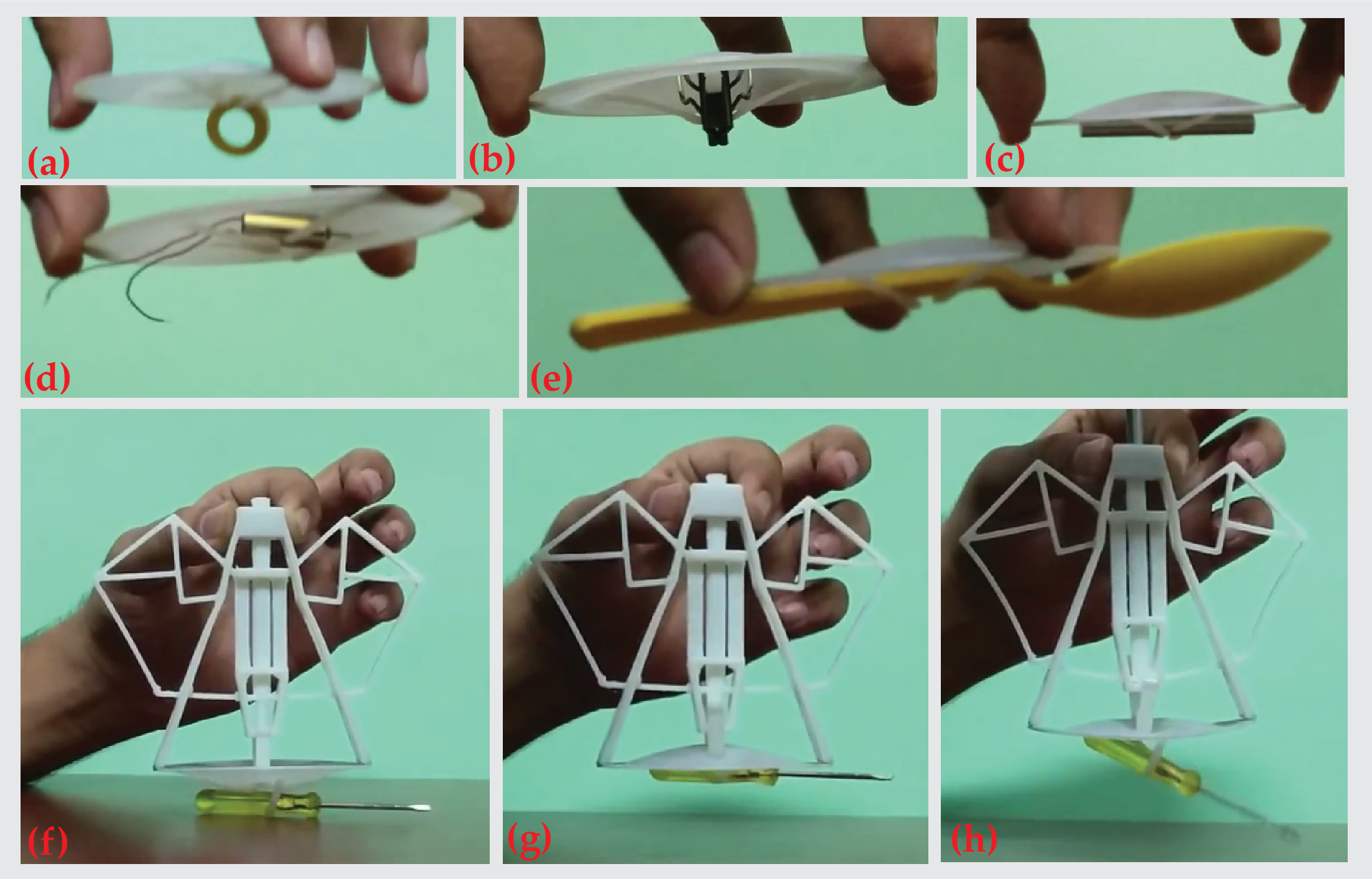}
	\caption{Grasping arms can capture objects of varying size, shape, and weight: (a) ring (b) paper clip (c) aluminum cylinder (d) mini motor (e) plastic spoon. 3D-printed prototype made in Object Conex 260. The grasping mechanism (f) grasps a screwdriver, (g) holds, and (h) releases it using the switching mechanism.}
	\label{fig:resultsAndPerformance}
\end{figure}

The 3D-printed prototype of the everting shell and grasping arms in Object Conex 260 is shown in \cref{fig:resultsAndPerformance}(a)-(e).
 The figure illustrates grasping of a range of objects such as: (a) ring (b) paper clip (c) aluminum cylinder (d) mini motor (e) plastic spoon. A complete 3D-printed prototype of the grasping mechanism is depicted in \cref{fig:resultsAndPerformance}(f)-(h). In \cref{fig:resultsAndPerformance}(f), gripper with an everted shell comes in contact with an screwdriver, which acts as the object here. The grasping mechanism holds the screwdriver by everting the shell in \cref{fig:resultsAndPerformance}(g). In \cref{fig:resultsAndPerformance}(h), the gripper releases the screwdriver by actuating at the actuating point of the switching mechanism.

One can note here that the gripper is able to grasp objects with length larger than the span of the planform of the shell; for example, the plastic spoon in \cref{fig:resultsAndPerformance}(e). This is possible due to the free space left in between the grasping arms by limiting their number to four. If one increases the number of grasping arms, the gripping performance improves but limits the size of graspable objects  due to the reduction in the free space between the grasping arms.

\section{Summary and future work}

A monolithic compliant grasping mechanism based on the bistability of an everting shell is proposed and illustrated with the aid of a 3D-printed prototype. The salient features of the grasping mechanism include lower power consumption, ability to grasp objects of a variety of  shapes, grasping of stiff objects initiated from contact,  and a completely monolithic design. Scope for future work in the design of everting shell, switching mechanism, and grasping arms is discussed next.

Cosine profile is not the only shape that shows bistability with fixed-fixed boundary conditions. For example, selecting the fundamental buckling mode of circular plate with fixed-fixed boundary conditions as the as-fabricated shape also shows bistability. Further, optimizing the arch-profile that fits into design constraints can improve the grasper functionality.

The gripper presented here holds objects that are stiff. However, grasping of delicate objects cannot be initiated from contact as it would applying excessive force on them. One possibility in holding such soft objects is to initiate the contraction of grasping arms actively, i.e., not from the contact with the object. This functionality could be achieved by redesigning the switching mechanism and the everting shell.

Rational design of the profile, orientation, cross-section parameters and the number of grasping arms can significantly improve the gripping performance. An optimal design that considers compliance, strength, and volume of the grasping arms would enhance the present design.   

\section*{Acknowledgement}
We acknowledge the grant from the Technology Initiative for Disabled and Elderly (TIDE) programme of the Department of Science and Technology, Government of India. 

\bibliographystyle{plain}
\bibliography{sample}

@article{krishnan2008evaluation,
	title={Evaluation and design of displacement-amplifying compliant mechanisms for sensor applications},
	author={Krishnan, Girish and Ananthasuresh, GK},
	journal={Journal of Mechanical Design},
	volume={130},
	number={10},
	pages={102304},
	year={2008},
	publisher={American Society of Mechanical Engineers}
}

@article{palathingal2017design,
	title={Design of bistable arches by determining critical points in the force-displacement characteristic},
	author={Palathingal, Safvan and Ananthasuresh, GK},
	journal={Mechanism and Machine Theory},
	volume={117},
	pages={175--188},
	year={2017},
	publisher={Elsevier}
}

@book {hibbett1998abaqus,
	title = {ABAQUS / standard: User's Manual},
	author = {Hibbett and Karlsson and Sorensen},
	volume = {1},
	year = {1998},
	publisher = {Hibbitt, Karlsson \& Sorensen}
}

@article{stavenuiter2017planar,
	title={A planar underactuated grasper with adjustable compliance},
	author={Stavenuiter, Ronald AJ and Birglen, Lionel and Herder, Just L},
	journal={Mechanism and Machine Theory},
	volume={112},
	pages={295--306},
	year={2017},
	publisher={Elsevier}
}

@article{laliberte2002underactuation,
	title={Underactuation in robotic grasping hands},
	author={Lalibert{\'e}, Thierry and Birglen, Lionel and Gosselin, Clement},
	journal={Machine Intelligence \& Robotic Control},
	volume={4},
	number={3},
	pages={1--11},
	year={2002}
}

@inproceedings{maruyama2013delicate,
	title={Delicate grasping by robotic gripper with incompressible fluid-based deformable fingertips},
	author={Maruyama, Ryoji and Watanabe, Tetsuyou and Uchida, Masahiro},
	booktitle={Intelligent Robots and Systems (IROS), 2013 IEEE/RSJ International Conference on},
	pages={5469--5474},
	year={2013},
	organization={IEEE}
}

@article{choi2006design,
	title={Design and feasibility tests of a flexible gripper based on inflatable rubber pockets},
	author={Choi, Ho and Koc, Muammer},
	journal={International Journal of Machine Tools and Manufacture},
	volume={46},
	number={12-13},
	pages={1350--1361},
	year={2006},
	publisher={Elsevier}
}

@article{manti2016stiffening,
	title={Stiffening in soft robotics: a review of the state of the art},
	author={Manti, Mariangela and Cacucciolo, Vito and Cianchetti, Matteo},
	journal={IEEE Robotics \& Automation Magazine},
	volume={23},
	number={3},
	pages={93--106},
	year={2016},
	publisher={IEEE}
}

@article{brown2010universal,
	title={Universal robotic gripper based on the jamming of granular material},
	author={Brown, Eric and Rodenberg, Nicholas and Amend, John and Mozeika, Annan and Steltz, Erik and Zakin, Mitchell R and Lipson, Hod and Jaeger, Heinrich M},
	journal={Proceedings of the National Academy of Sciences},
	volume={107},
	number={44},
	pages={18809--18814},
	year={2010},
	publisher={National Acad Sciences}
}

@article{amend2012positive,
	title={A positive pressure universal gripper based on the jamming of granular material},
	author={Amend, John R and Brown, Eric and Rodenberg, Nicholas and Jaeger, Heinrich M and Lipson, Hod},
	journal={IEEE Transactions on Robotics},
	volume={28},
	number={2},
	pages={341--350},
	year={2012},
	publisher={IEEE}
}

@book{murray2017mathematical,
	title={A mathematical introduction to robotic manipulation},
	author={Murray, Richard M},
	year={2017},
	publisher={CRC press}
}

@inproceedings{bicchi2000robotic,
	title={Robotic grasping and contact: A review},
	author={Bicchi, Antonio and Kumar, Vijay},
	booktitle={ICRA},
	volume={348},
	pages={353},
	year={2000},
	organization={Citeseer}
}

@inproceedings{nguyen2016gripper,
	title={A gripper based on a compliant bitable mechanism for gripping and active release of objects},
	author={Nguyen, Thang-An and Wang, Dung-An},
	booktitle={Manipulation, Automation and Robotics at Small Scales (MARSS), International Conference on},
	pages={1--4},
	year={2016},
	organization={IEEE}
}

\vskip2pc

\end{document}